\documentclass[twoside]{article}

\usepackage{lipsum} 

\usepackage[sc]{mathpazo} 
\usepackage[T1]{fontenc} 
\usepackage{microtype} 

\usepackage[hmarginratio=1:1,top=32mm,columnsep=20pt]{geometry} 
\usepackage{multicol} 
\usepackage[hang, small,labelfont=bf,up,textfont=it,up]{caption} 
\usepackage{booktabs} 
\usepackage{float} 
\usepackage{hyperref} 

\usepackage{lettrine} 
\usepackage{paralist} 

\usepackage{abstract} 

\usepackage{titlesec} 
\renewcommand\thesection{\Roman{section}} 
\renewcommand\thesubsection{\Roman{subsection}} 
\titleformat{\section}[block]{\large\scshape\centering}{\thesection.}{1em}{} 
\titleformat{\subsection}[block]{\large}{\thesubsection.}{1em}{} 

\usepackage{fancyhdr} 
\usepackage{graphicx}
\usepackage{caption}
\usepackage{subcaption}
\usepackage{hyperref}
\hypersetup{
colorlinks,
citecolor=black,
filecolor=black,
linkcolor=black,
urlcolor=black
}
\usepackage{color}

\title{\vspace{-15mm}\fontsize{24pt}{10pt}\selectfont\textbf{Text authorship identified using the dynamics of word co-occurrence networks}}

\author{
\large
\textsc{Camilo Akimushkin$^{1}$ }\\
\textsc{Diego Raphael Amancio$^{2}$ }\\
\textsc{Osvaldo N. Oliveira Jr.$^{1}$\footnote{Corresponding author: \href{mailto:chu@ifsc.usp.br}{chu@ifsc.usp.br}} }\\
\normalsize $^{1}$S\~ao Carlos Institute of Physics, University of S\~ao Paulo, S\~ao Carlos, S\~ao Paulo, Brazil \\
\normalsize $^{2}$Institute of Mathematics and Computer Science, University of S\~ao Paulo, S\~ao Carlos, S\~ao Paulo, Brazil \\
\vspace{-5mm}
}
\date{}

\begin{document}

\maketitle 

\thispagestyle{empty} 


\begin{abstract}

The identification of authorship in disputed documents still requires human expertise, which is now unfeasible for many tasks owing to the large volumes of text and authors in practical applications. In this study, we introduce a methodology based on the dynamics of word co-occurrence networks representing written texts to classify a corpus of 80 texts by 8 authors. The texts were divided into sections with equal number of linguistic tokens, from which time series were created for 12 topological metrics. The series were proven to be stationary (p-value>0.05), which permits to use distribution moments as learning attributes. With an optimized supervised learning procedure using a Radial Basis Function Network, 68 out of 80 texts were correctly classified, i.e. a remarkable 85\% author matching success rate. Therefore, fluctuations in purely dynamic network metrics were found to characterize authorship, thus opening the way for the description of texts in terms of small evolving networks. Moreover, the approach introduced allows for comparison of texts with diverse characteristics in a simple, fast fashion.
\end{abstract}

\textbf{Keywords:} Complex networks; network dynamics; text analysis; authorship identification.


\section*{Introduction}

Statistical methods have long been applied to analyze written texts and language patterns~\cite{zipf1935psycho}, which now include network representations of text to investigate linguistic phenomena~\cite{ferrer2001two,i2001small,concentrico,diegohybrid,PhysRevE.83.026103,wikinfo,PhysRevE.91.032810}. Networks generated from text share several features with other complex systems, e.g. transportation networks~\cite{stretMas}, biological systems~\cite{importantNodes,kaiser}, social interactions~\cite{PhysRevE.68.036122}. Examples of language-related networks include phonological networks with modular or cut-off power-law behaviors~\cite{kapatsinski2006sound,mukherjee2007modeling,mukherjee2009self}, semantic similarity networks with small-world and scale-free properties~\cite{sigman2002global}, syntactic dependency networks with hierarchical and small-world organization~\cite{ferrer2014stronger,i2004patterns} and collocation networks, which also display small-world and scale-free properties~\cite{i2001small}. The ubiquity of specific patterns in language networks is believed to account for an easy navigation and acquisition in semantic and syntactic networks~\cite{ontogeny}.
Of particular relevance to this study, word co-occurrence networks are a special case of collocation networks where two words (nodes) are linked if they appear close to each other in a text. Co-occurrence networks are convenient because they do not require prior linguistic knowledge, apart from that needed to filter relevant information. Since most of the syntactic relations occur between adjacent words, co-occurrence networks can be seen as simplified versions of syntactic networks~\cite{i2004patterns}. Several patterns have been identified in co-occurrence networks formed from large corpora, such as the power-law regimes for degrees distribution~\cite{ferrer2001two} and core-periphery structure~\cite{choudhury2010global} resulting from the complex organization of the lexicon. The overall structure and dynamics of networks representing texts have been modeled to describe their mechanism of growth and attachment~\cite{barabasi1999emergence,dorogovtsev2001language}, while nuances in the topology of real networks were exploited in practical problems, including natural language processing ~\cite{biemann2009networks,amancio2012identification,amancio2011comparing}. In this study, we use the co-occurrence representation to probe how the variation of network topology along a text is able to identify an author's style.

Writing style is more subjective than other text characteristics (e.g. topic), making authorship recognition one of the most challenging text mining tasks~\cite{stamatatos2009survey,basile2008example}. It is crucial for practical applications such as text classification~\cite{amancio2011comparing}, copyright resolution~\cite{chaski2005s}, identification of terrorist messages~\cite{Abbasi} and of plagiarism~\cite{stamatatos2009survey}. Early studies using stylometry were conducted by Mosteller and Wallace to identify authorship of the so-called Federalist Papers~\cite{mosteller1964inference}. A myriad of methods to tackle the problem have been developed since then, typically using statistical properties of words (e.g. mean length, frequency, burstiness and vocabulary richness) and characters (e.g. character counts and long-range correlations) ~\cite{stamatatos2009survey}, in addition to syntactic and semantic information~\cite{stamatatos2009survey}. Methods from statistical physics have also been used for authorship recognition~\cite{Havlin1995148,Pomi}, which in recent years included text modeling with co-occurrence networks~\cite{Liang20094901,amancio2012complex,disentangling,liu2013language,segarra,10.1371/journal.pone.0118394}. The adequacy of co-occurrence networks for the task was confirmed for the first time with the correlation between network topology and authors' styles~\cite{amancio2011comparing}. Despite this relative success, some issues concerning the applicability of network methods remain unsolved.

A major issue in network representation is that regular patterns among concepts only emerge when large pieces of texts are available. Furthermore, rigorous network-based similarity estimators usually assume that the networks comprise the same number of nodes and edges, since most measurements are affected by the network size~\cite{newman2010introduction}. Unfortunately, such strong assumption often does not hold for written texts ranging from small tales to long novels, which may hinder the applicability of models to real situations. As we shall show, the method presented here obviates these issues with a simple approach based on network dynamics.

\section*{Methods}

Written texts were represented as co-occurrence networks, from which a set of network dynamics measurements were obtained. These measurements were used as attributes in pattern recognition methods in order to identify the author of a given text. The construction and analysis of the measurements are described in detail in the following subsections.

\subsection*{Modeling texts as co-occurrence networks}

The texts used for classification come from a collection of novels and tales in English whose details are provided in the Supporting Information. The collection comprising $8$ authors with $10$ texts per author was selected to simulate a real case where the text lengths are varied in a range from $2,853$ to $267,012$ tokens with an average of $53,532$ tokens.

The approach introduced requires a pre-processing step before transforming texts into networks, which consists in the removal of stopwords and lemmatization. Because we are mostly interested in the relationship between content words, stopwords such as function words conveying low semantic information were removed as in many studies of this type~\cite{cong2014approaching}. The remaining words were lemmatized so that nouns and verbs were mapped to their singular and infinitive forms, and therefore words related to the same concept were mapped into the same node (also referred to as one single token). Since lemmatization requires part-of-speech (POS) tagging, we used the maximum-entropy approach described in~\cite{GreRub}. The co-occurrence networks were constructed with each distinct word becoming a node and two words being linked if they were adjacent in the pre-processed text~\cite{amancio2011comparing}. The link is directed from the word appearing first to the second word and is weighted by the number of times the pair is found in the text.

\subsection*{Characterization of written texts via network dynamics}

The proposed method for authorship attribution is based on the evolution of the topology of networks, i.e. we exploit the network dynamics. Therefore, unlike previous approaches (see e.g.~\cite{i2001small,concentrico,roxas2010prose}), we do not construct one single network from the whole book. Instead, a piece of text is divided into shorter parts comprising the same number of tokens. Then, a co-occurrence network is constructed for each part, which generates a series of independent networks for each book. The last partition is disregarded from the analysis because it is shorter than the previous ones. Since distinct books have different numbers of tokens, the series length varies from book to book.

Each partition is described by the following topological network measurements:
clustering coefficient $C$, which gives the fraction of possible triangles that exist for a particular node;
network diameter $D$, which is the largest of all shortest paths ($\mathrm{max}\{S_{ij}\}$);
network radius $R$, which is the smallest of all shortest paths ($\mathrm{min}\{S_{ij}\}$);
number of cliques (complete subgraphs) $C_q$;
load centrality $L$, similar to betweenness centrality but considering weights on edges;
network transitivity $T$, which measures the fraction of all connected triples which are in fact triangles, $T=3\times \mathrm{triangles}/\mathrm{triads}$;
betweenness centrality $B$, which measures how many shortest paths pass through a given node;
shortest path length $S$, which is the shortest number of edges between two nodes;
degree $K$ or connectivity (number of edges) of a node;
intermittency $I$, which measures how periodically a word is repeated~\cite{berryman2003statistical};
total number of nodes $N$ (i.e. vocabulary size);
and total number of edges $E$.
Even though intermittency is not a traditional network measurement, we considered it because of its strong relationship with the concept of cycle length in networks. Moreover, this measurement has been proven useful for analyzing text styles~\cite{amancio2011comparing}. The metrics $D$, $R$, $C_q$, $T$, $N$ and $E$ are scalar values for a network, while the other measurements are computed for each node individually. In order to have an overall picture of each partition, we computed the average values of $C$, $L$, $B$, $S$, $K$ and $I$. As such, each partition is characterized by a set of twelve global topological measurements.

The total number of tokens $W$ (equal to the total weight among links), in each partition, was selected in a simple optimization procedure, with a compromise between having a long but noisy series (many small networks) and a shorter, more stable one (few large networks). We found that with $W=200$ tokens one ensures a series length with $T=268$ elements on average while keeping the number of nodes over $n=100$ for all networks.

\begin{center}
\begin{figure}[t]
\includegraphics[width=0.97\columnwidth]{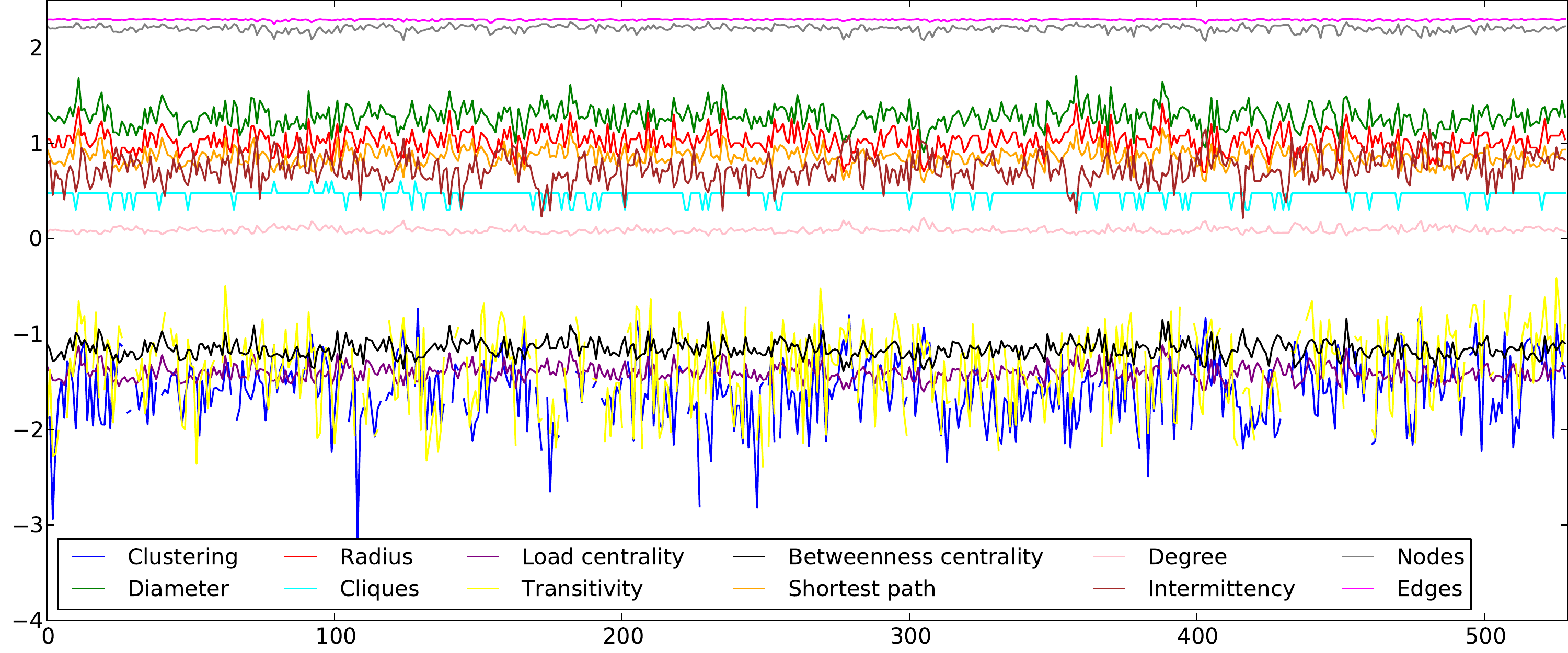}
\caption{Time series for Moby Dick by Herman Melville. The horizontal axis denotes the index of realizations, and the vertical axis brings the base $10$ logarithm of the metrics identified in the inset.}
\label{fig:ts}
\end{figure}
\end{center}

\begin{center}
\begin{figure}[t]
\begin{tabular}{cc}
\includegraphics[width=0.57\columnwidth]{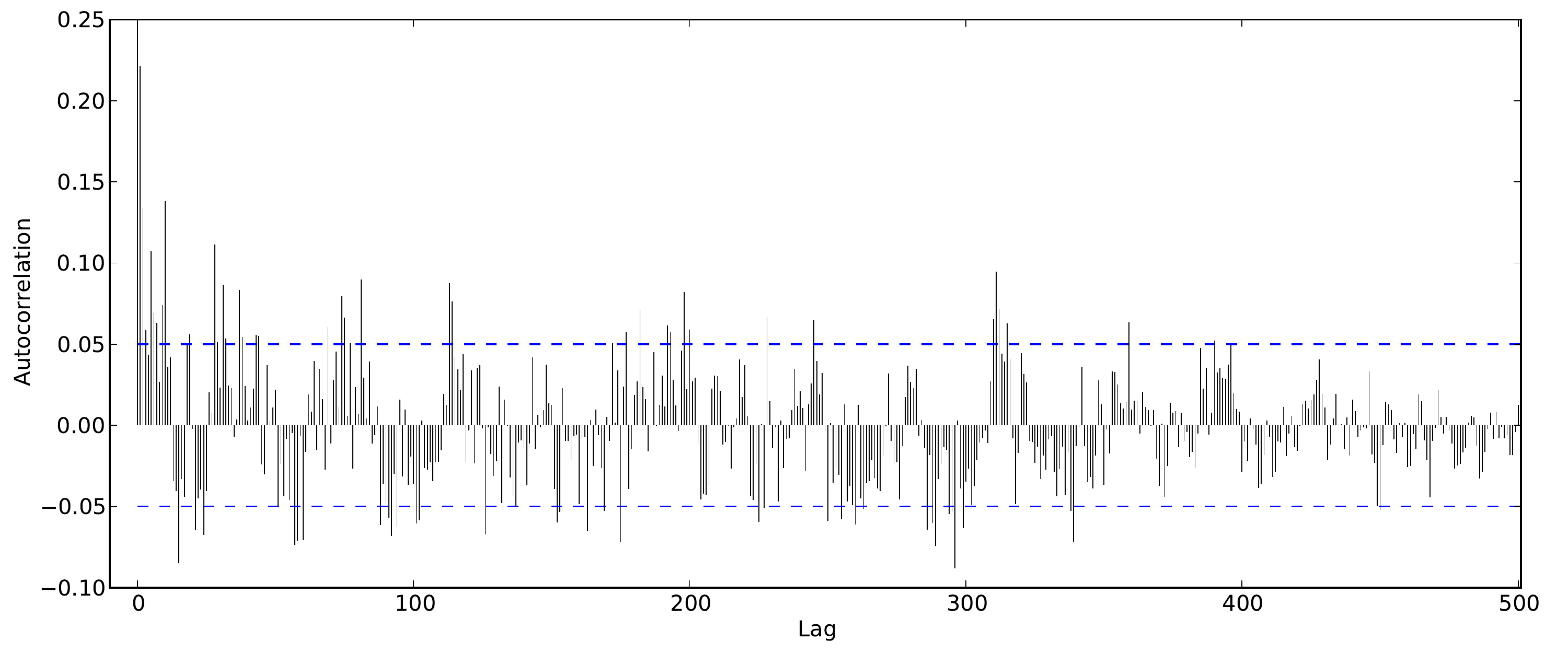} &
\includegraphics[width=0.38\columnwidth]{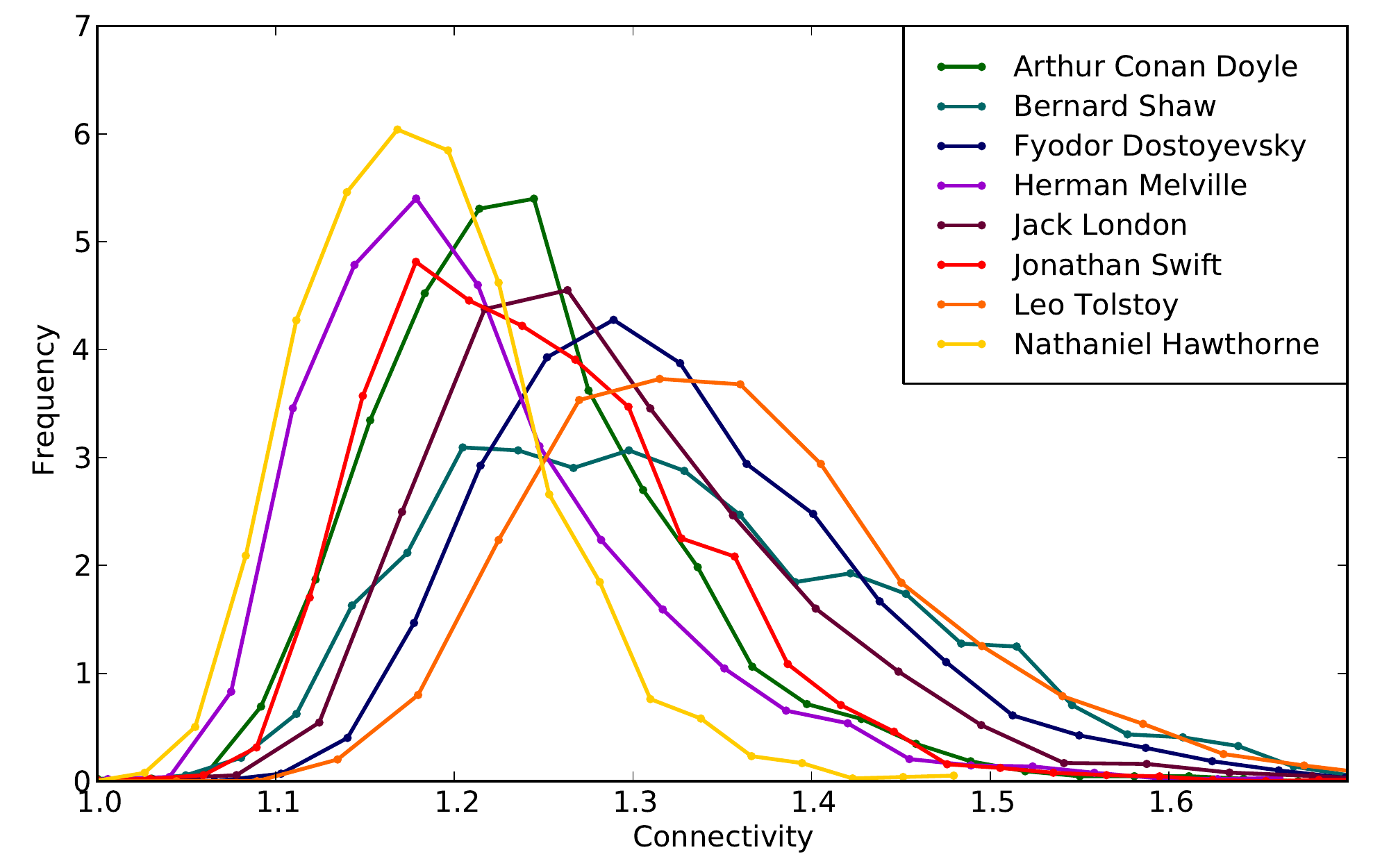}\\
(a) Autocorrelation &
(b) Histograms
\end{tabular}
\caption{(a) Autocorrelation for the series of clustering coefficient of Moby Dick by Herman Melville. Dashed lines mark the $5\%$ threshold which are surpassed only by chance. (b) Histograms for time series of degree $K$ (connectivity) from all books on the collection by author. The distributions have characteristic moments for each author.}
\label{fig:auto+series_hist}
\end{figure}
\end{center}

A set of time series is constructed by extracting the twelve global network metrics defined above for each of the networks from a book. Figure \ref{fig:ts} shows the series for Moby Dick by Herman Melville, from which one may note that the series oscillate steadily around a fixed value along the text, with no significant trend. Indeed, the analysis is facilitated if the series are stationary. Strong stationarity requires the expected values being constant along time while weak stationarity implies that the mean value (and sometimes the variance) is constant.
We confirmed that the time series are stationary, i.e. characterized by low values of autocorrelation. Correlation of a time series with itself shifted by a certain gap measures how much a value in the series depends on the previous ones, implying that the autocorrelation must be almost null for all but the first few values of the gap.
In order to assess series stationarity, we implemented Kwiatkowski-Phillips-Schmidt-Shin (KPSS), augmented Dickey-Fuller, and MacKinnon (finite-sample and asymptotic) unit root tests~\cite{kwiatkowski1992testing,said1984testing,mackinnon1996numerical}. These tests assume stationarity as the null hypothesis. We observed that for ten of the twelve time series, $p$-values were larger than $0.05$, and therefore the stationarity hypothesis cannot be rejected.
For the two remaining series (clustering coefficient $C$ and transitivity $T$) two of the four $p$-values were less than $0.05$: those for augmented Dickey-Fuller and finite-sample MacKinnon tests. MacKinnon~\cite{mackinnon1996numerical} pointed out that $z$-tests can lead to unreliable finite-sample $p$-values substantially smaller than asymptotic $p$-values as we have observed for the clustering coefficient and transitivity. Moreover, autocorrelation drops after a small lag as shown in figure~\ref{fig:auto+series_hist}(a) for the series of clustering coefficient which can be fitted with an ARIMA(1,1,2) model, thus implying that the first derivative of the series is stationary.

The finding that the series can be considered stationary allows one to compare estimated values from series of different lengths.
As the distribution in the time series was found to display a bell-shaped form (shown in figure~\ref{fig:auto+series_hist}(b)), we propose the first four moments of the series distributions as the dynamical measurements, i.e.
\begin{equation}
     \mu_i=\left[ \frac{1}{T-1} \sum_{j=1}^T (x_j-\mu_1)^i \right]^{1/i},
\end{equation}
where $1<i\leq 4$, $x_j$ are the series values and $\mu_1$ is the average of the measurements in the series. Since there are twelve time series, we obtain $48$ dynamical measurements to characterize a book.

\subsection*{Data analysis}

The moments of the network metrics are used to characterize a text. In the terminology of machine learning these are the attributes (also called features) for the algorithms, while individual texts are the instances and the author of a text corresponds to the instance class. A $80\times48$ data matrix is constructed where each row corresponds to an instance and each column corresponds to an attribute. In order to account for the different scales of the attributes (see figure \ref{fig:ts}), each column in the data matrix is normalized between zero and one. From the data matrix, the author of each book is inferred using standard supervised learning (classification) algorithms~\cite{witten2005data}.

There is a dimensionality reduction stage prior to the classification. Dimensionality reduction is achieved through either feature selection or feature extraction. Feature selection consists in removing attributes which do not satisfy a given condition, thus leaving a subset or combination of the total number of features. On the other hand, feature extraction blends the original attributes together in order to create a set of, usually fewer, new attributes. Feature selection was implemented using variance threshold and scoring criteria. Variance threshold selection imposes a minimum variance among the realizations of an attribute, for example, if an attribute has the same value for all instances its variance is zero and can be safely removed because it does not contribute to the classification process. We also implemented feature selection based on score. The huge number of combinations of attributes prohibits an exhaustive search of the combination(s) with the highest score. Instead, we start by testing all subsets obtained by removing one attribute from the whole set. In the next step we test all subsets obtained by removing one attribute from the subsets with the highest score in the previous step, and the process is iterated. Dimensionality reduction through feature extraction was implemented using the well-known Principal Component Analysis (PCA) and Isomap~\cite{tenenbaum2000global,paulovich2007projection} technique. Isomap analyzes data points in a high-dimensional space that are implicitly located in a curved manifold of smaller dimensionality. Dimensionality reduction is then achieved by unwrapping the manifold and getting rid of the extra dimensions. As will be shown, both feature selection and extraction improve the classification success score.

Since there are many supervised learning algorithms, we have selected some to cover the most distinct classification paradigms: ZeroR (0R), which arbitrarily labels all instances as belonging to the most prevalent class; OneR (1R), which ranks attributes according to their error rate; Naive Bayes (NB), which assumes independence among attributes; K-nearest neighbors (KNN), where the class of an instance is inferred by a voting process over the nearest neighbors in the training dataset; J$48$, which organizes the patterns in a tree-like structure; and Radial Basis Function Network (RBFN) where a learning network with an input, a processing, and an output layer is used. Due to their simplicity, 0R and 1R are only used for comparison. In all methods, the parameters were set with their default configuration~\cite{amancio2014systematic} and the classification is calculated for a 10-fold stratified cross-validation. A detailed description of classification algorithms can be found in~\cite{witten2005data}.

The performance of supervised and unsupervised algorithms can be evaluated with two standard scores: precision and recall. Both are real values ranging from zero to one, being specific for a given class $c$. Precision ($P_c$) is defined as
\begin{equation}
P_c = \frac{\tau_c}{\tau_c+\epsilon_c},
\label{eq:precision}
\end{equation}
where $\tau_c$ is the number of instances belonging to class $c$ that were correctly classified (i.e. the number of true positives), and $\epsilon_c$ is the number of instances of other classes that were wrongly classified as belonging to class $c$ (i.e. number of false positives).
The Recall $R$ for class $c$ is computed as
\begin{equation}
R_c = \frac{\tau_c}{\tau_c+\gamma_c},
\label{eq:recall}
\end{equation}
where $\gamma_c$ is the number of instances of class $c$ that were incorrectly classified (i.e. the number of false negatives).
The precision and recall scores defined above refer to a single class. To obtain a single value from the dataset, one may use micro- and macro-averaging. Micro-averaging weights the scores of each class by the number of instances; therefore, classes with many instances are more relevant to the overall score. In contrast, the macro-average score is the arithmetic mean of the scores of all classes. Note that the micro-averaged recall is equivalent to the success rate, that is, the proportion between the number of instances correctly classified and the total number of instances. For the present collection, having the same number of instances per class, micro- and macro-averaging are equivalent.

\section*{Results and Discussion}

The authorship signature is captured by the metrics proposed, which reveals the relationship between style and changes in network structure. Success scores greatly surpass the threshold imposed by a blind classification obtained with ZeroR algorithm, which for our collection is $1/8=12.5\%$. Unmodified data from the $48$ original metrics were classified with success rates in the range from $45\%$ to $62.5\%$ as shown in table~\ref{table:scores}. The simple OneR algorithm also performed well, reaching $46.25\%$ score.

\begin{table}
\caption{Summary of success scores for the whole set of $48$ attributes and subsets after feature selection. The results for combinations of attributes which obtained the best scores under variance threshold and score-based criteria are presented along with those for the subset of the first moments $\{\mu_1\}$ and the complementary subset of all higher moments $\{\mu_1,\mu_2,\mu_3\}$.}
\label{table:scores}
\begin{center}
\begin{tabular}{lllll}
Combination				& J48 $(\%)$	& KNN $(\%)$	& NB $(\%)$	& RBFN $(\%)$	\\
\hline
Whole set				& $45$			& $62.5$		& $61.25$	& $56.25$ \\
Variance threshold best	& $55$			& $67.5$		& $63.75$	& $63.75$ \\
Score-based best		& $75$			& $78.75$		& $77.5$	& $75$ \\
$\{\mu_1\}$				& $45$			& $43.75$		& $46.25$	& $40$ \\
$\{\mu_1,\mu_2,\mu_3\}$	& $38.75$		& $63.75$		& $60$		& $57.5$ \\
\hline
\end{tabular}
\end{center}
\end{table}

\begin{figure}[t]
    \centering
    \begin{subfigure}[b]{0.48\textwidth}
        \includegraphics[width=\textwidth]{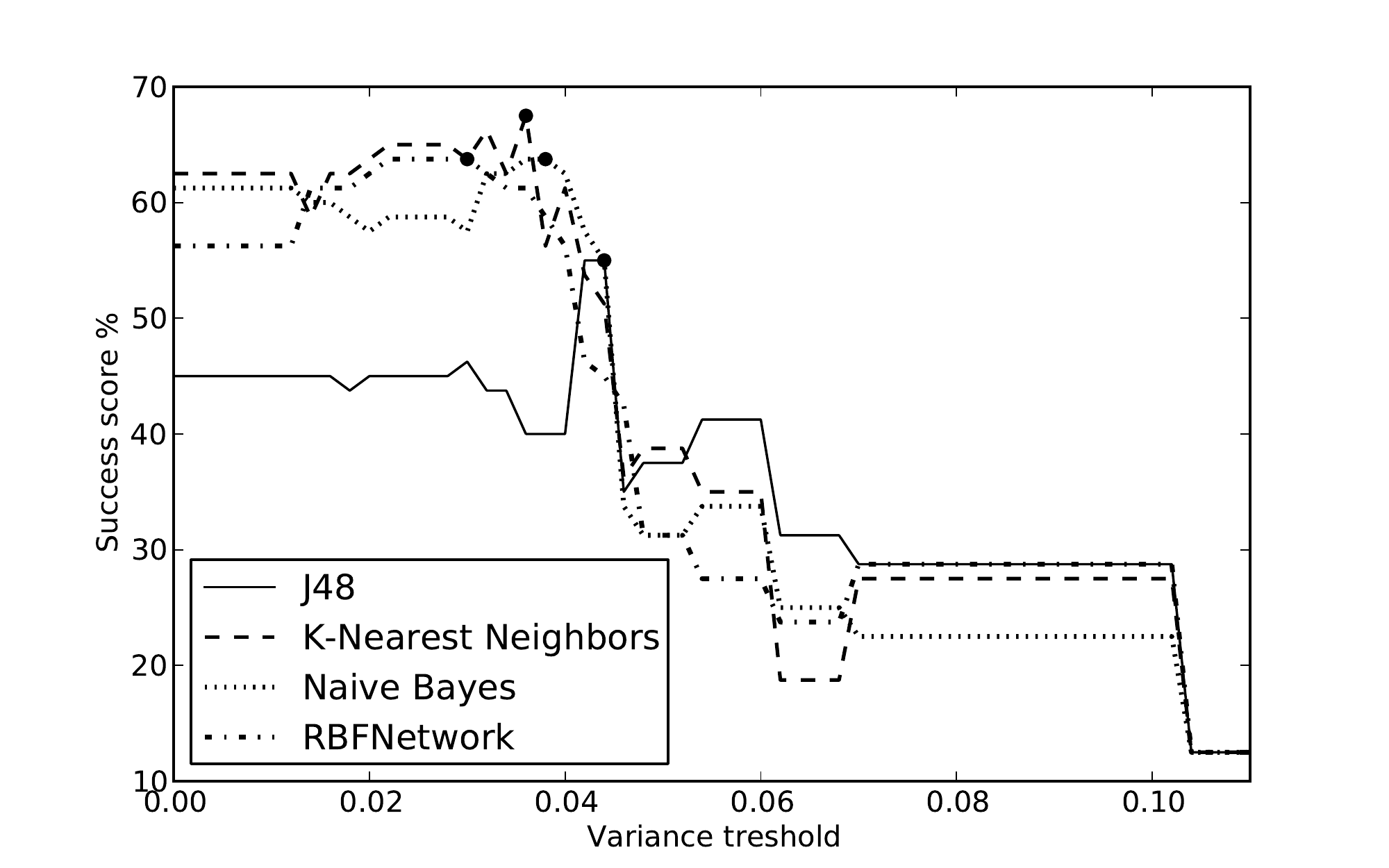}
        \caption[a]{Variance threshold scores.}
        \label{fig:vts}
    \end{subfigure}
    ~ 
    \begin{subfigure}[b]{0.48\textwidth}
        \includegraphics[width=\textwidth]{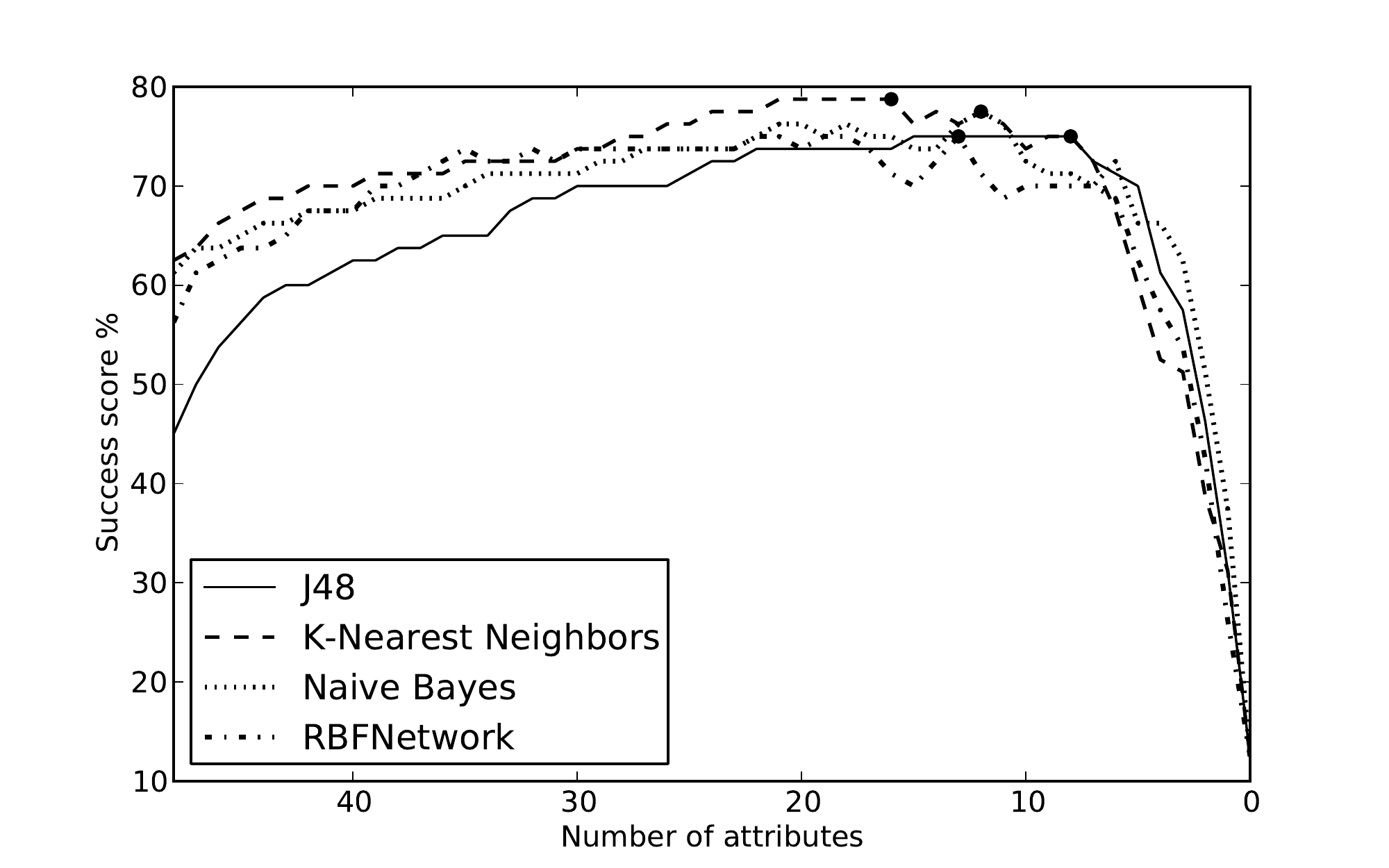}
        \caption[c]{Score-based scores.}
        \label{fig:sbs}
    \end{subfigure}

    \begin{subfigure}[b]{0.48\textwidth}
        \includegraphics[width=\textwidth]{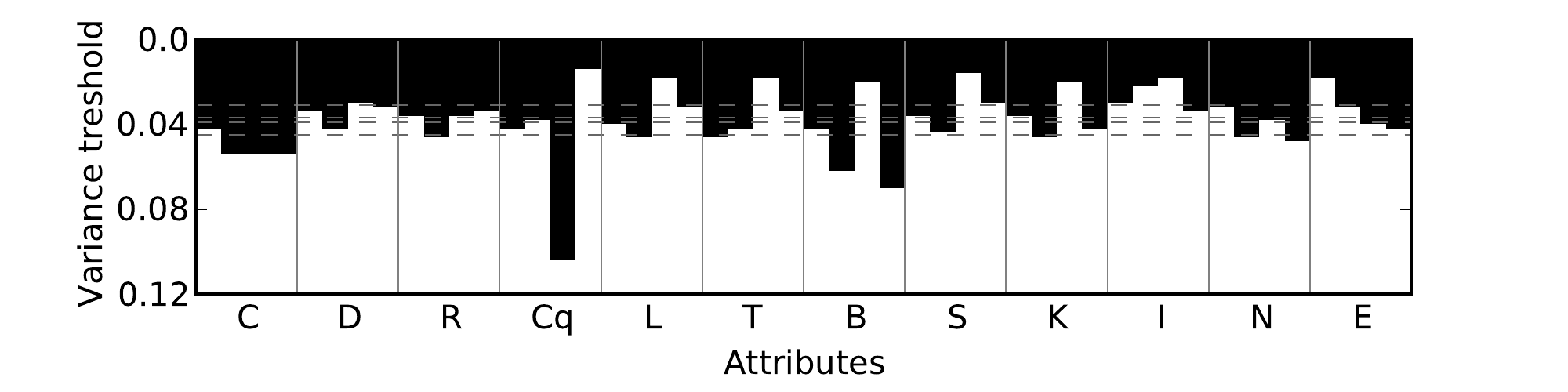}
        \caption[b]{Variance threshold attributes.}
        \label{fig:vta}
    \end{subfigure}
    ~ 
    \begin{subfigure}[b]{0.48\textwidth}
        \includegraphics[width=\textwidth]{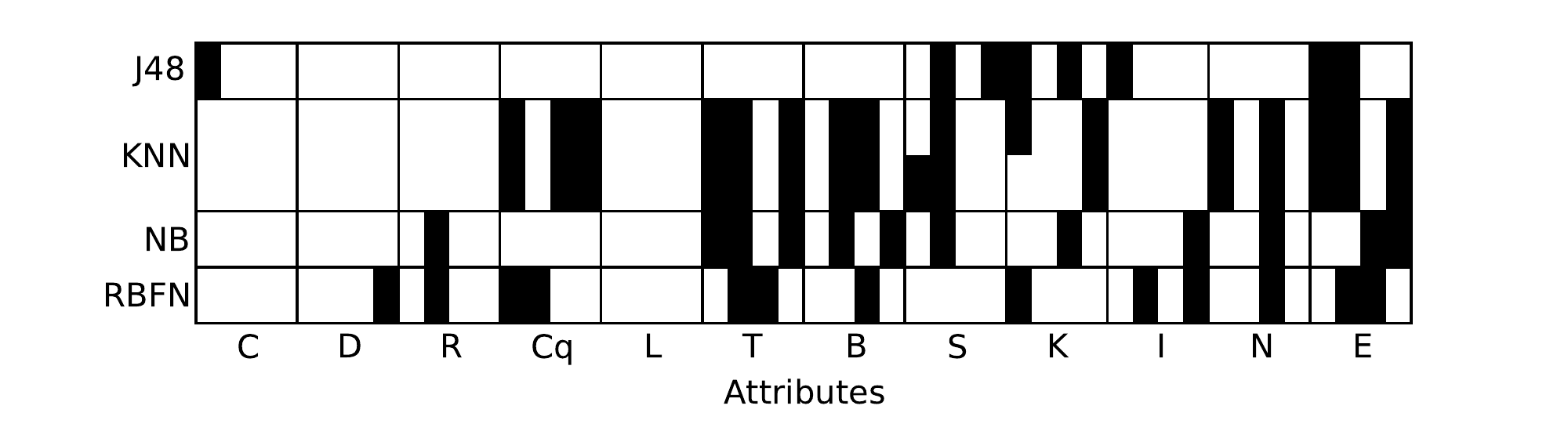}
        \caption[d]{Score-based attributes.}
        \label{fig:sba}
    \end{subfigure}
    \caption{Success scores and combinations of attributes using the variance threshold ((a) and (c)), and using score-based feature selection ((b) and (d)). In (a), (b) the maximum values with minimum number of attributes are marked with circles. In (c), (d) for each network metric (represented by a label in the horizontal axis) the four first moments are presented in increasing order from left to right. A black cell indicates that the attribute is present in the combination. For the variance threshold feature selection, there is a unique combination that satisfies every threshold marked by the vertical axis in (c). For instance, thresholds for the four maximum scores in (a) are marked in (c) by the four dashed horizontal lines. For the score-based feature selection there can be multiple combinations of attributes with the same number of attributes and the same score. Only the combinations with maximum scores and marked with circles in (b) are presented in (d); for KNN algorithm there were two combinations with maximum score.}
\label{fig:feat_sel}
\end{figure}

Dimensionality reduction using either feature extraction or feature selection increased the success rates for all algorithms. The results of feature selection are shown in figure~\ref{fig:feat_sel} for both variance threshold and score-based selection. In both cases the best results are obtained with an intermediate number of metrics: we begin by removing misleading attributes, therefore improving classification; however, at the end of the process most of the attributes which carry important information are removed and then the classification scores are lowered. In figures~\ref{fig:feat_sel}(a) and~\ref{fig:feat_sel}(b) success scores are presented, with the maximum value for each curve marked with a circle. If there is more than one maximum (e.g. J48 and NB for variance threshold and J48 and KNN for score-based selection), we only consider the combinations with the fewest number of attributes, located at the rightmost positions.

The results of feature selection using a variance threshold are shown in figures~\ref{fig:feat_sel}(a) and~\ref{fig:feat_sel}(c). There is a single subset (combination) of attributes for each variance threshold level. At the lowest threshold in figure~\ref{fig:feat_sel}(c) all attributes are present (even though the threshold is larger than zero) and all cells of the highest row are colored black. As the threshold is gradually increased, attributes are successively removed until there are no attributes left and all the cells in the lowest row are colored white. Remarkably, the first and the last attributes removed were respectively the fourth and the third moments of the number of cliques $C_q$. Note also that for nine of the twelve network metrics, either the third or the fourth moment had the smallest variance. The maximum scores are marked with circles in figure~\ref{fig:feat_sel}(a) and listed in table~\ref{table:scores}. The thresholds for maximum scores marked in~\ref{fig:feat_sel}(a) are located in a narrow range and are represented in figure~\ref{fig:feat_sel}(c) as dashed lines.

The results of feature selection based on score are shown in figures~\ref{fig:feat_sel}(b) and~\ref{fig:feat_sel}(d). We start with all the attributes in the left end of figure~\ref{fig:feat_sel}(b). As we explore the combinations obtained by removing one attribute at a time the scores increase (monotonically for J48 and KNN) until a maximum value is reached, after which the scores rapidly decrease reaching ZeroR score when there are zero attributes. The maximum scores are marked with circles in figure~\ref{fig:feat_sel}(b) and listed in table~\ref{table:scores}. It must be noted that the maximum scores can be reached with a few attributes, at most $16$ attributes in the case of KNN. The combinations of attributes giving the maximum scores marked are presented in detail in figure~\ref{fig:feat_sel}(d). For KNN two combinations of attributes reached the highest score. Again, the four moments of a given network metric are grouped together. It can be seen that the best scoring combinations for some algorithms did not include any of the four moments from some network metrics. In particular, load centrality $L$ was not used by any algorithm (having therefore a blank column for $L$ in figure~\ref{fig:feat_sel}(d)). One should highlight the betweenness centrality $B$, which was extensively used by KNN, NB and RBFN even though its mean value (i.e. first moment, and the leftmost column under the $B$ label on figure~\ref{fig:feat_sel}(d)) was not used by these algorithms.

Two last combinations of attributes were constructed. The first moments $\mu_1$ represent the static metrics previously studied (see e.g. ~\cite{amancio2011comparing,mehri2012complex}) and define a subset of $12$ attributes. The complementary subset of $36$ second, third, and fourth moments represent the dynamical aspects of networks since they describe the extent of variation around the mean value throughout a text. Classification was applied to these two subsets without further dimensionality reduction. The results are listed in the fourth and fifth rows of table~\ref{table:scores} showing that purely dynamical metrics provide better overall performance when compared to the statical counterparts, while both subsets score similarly to the whole set of $48$ attributes.

Another dimensionality reduction technique implemented was feature extraction, using both linear PCA and nonlinear Isomap. The latter uses geodesic distances in an embedded manifold instead of high-dimensional Euclidean distances. There is a free parameter in Isomap: the number of neighbors $n_{neighbors}$. The distance between two instances considered neighbors is the traditional Euclidean distance while the distance between two farthest instances is the geodesic distance for a path inside the manifold~\cite{tenenbaum2000global}. The results for Isomap depend on $n_{neighbors}$ and on the reduced number of dimensions $n_{comps}$; we varied both parameters from $2$ to $15$ and found similar results for most cases (see Supporting Information). The best scores reported below were obtained for $n_{neighbors}=10$ and $n_{comps}=13$.

\begin{figure}[t]
    \centering
    \begin{subfigure}[b]{0.48\textwidth}
        \includegraphics[width=\textwidth]{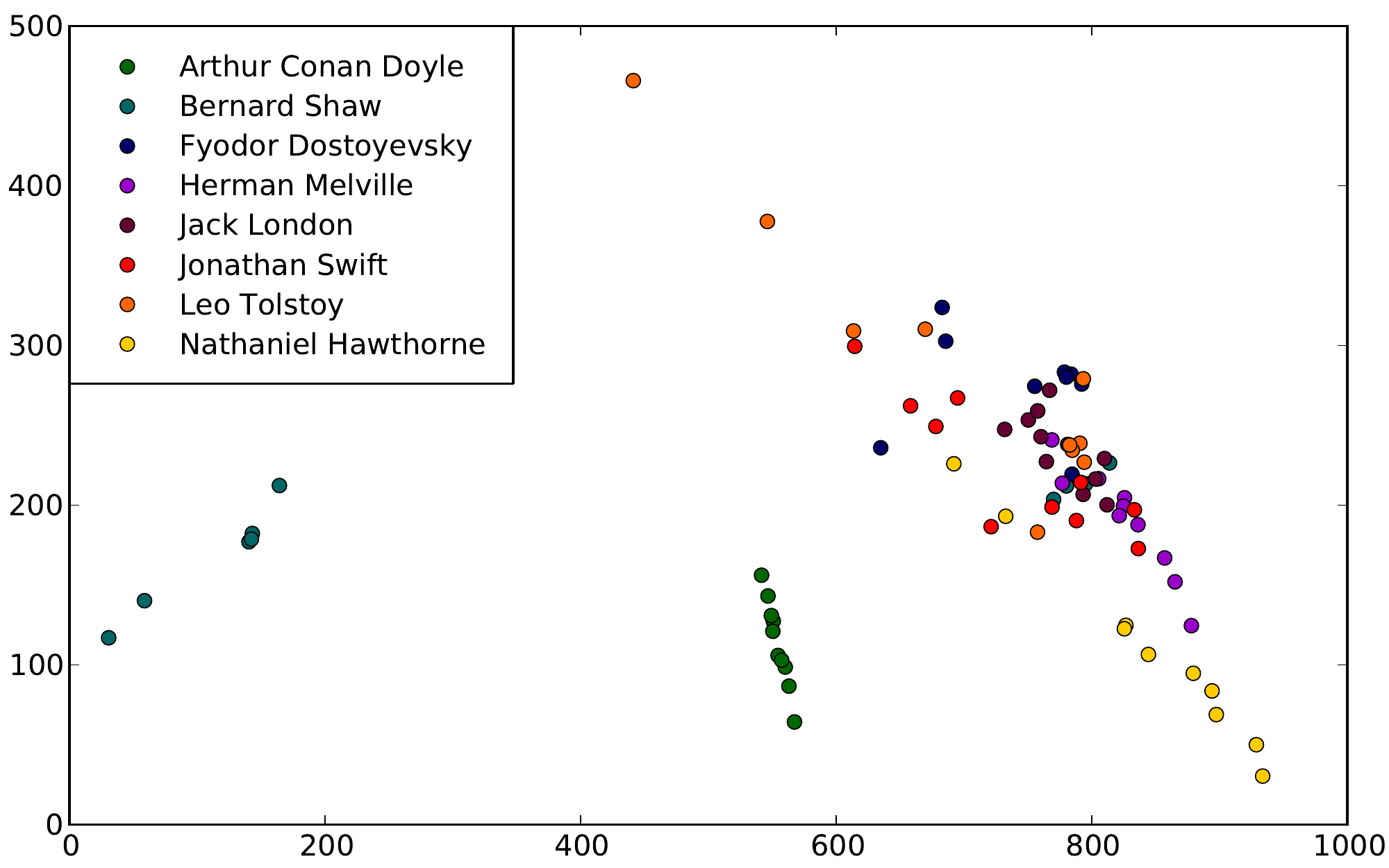}
        \caption{Isomap attribute space.}
        \label{fig:feat_ext_isomap}
    \end{subfigure}
    ~ 
    \begin{subfigure}[b]{0.48\textwidth}
        \includegraphics[width=\textwidth]{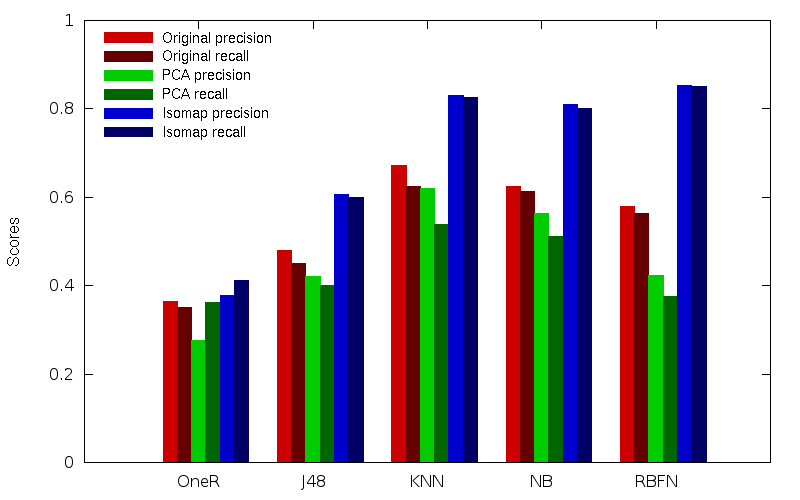}
        \caption{Classification validation with feature extraction.}
        \label{fig:feat_ext_scores}
    \end{subfigure}
    \caption{(a) First two dimensions of attribute space after Isomap. Each point represents a book and each color represents an author. (b) Validation of the classification without dimensionality reduction (red), and with feature extraction using PCA (green) and Isomap (blue).}
\label{fig:feat_ext}
\end{figure}

Figure~\ref{fig:feat_ext}(b) shows precision (defined by equation~\ref{eq:precision}) and recall (success score, defined by equation~\ref{eq:recall}) for original (without dimensionality reduction), PCA-, and Isomap-treated attributes. Dimensionality reduction through PCA leads to lower precision and recall, while Isomap enhances the classification efficacy of the algorithms. The best performance is reached with RBFN for which the authorship of $68$ of the $80$ texts in the collection is correctly identified, thus reaching $85\%$ success score (recall) and $0.854$ precision. This performance is robust among algorithms as both precision and recall surpass $80\%$ using KNN, NB and RBFN. For visualization purposes Isomap was also applied to reduce the number of attributes to a two-dimensional space using the Projection Explorer software \cite{paulovich2007projection} as shown in figure~\ref{fig:feat_ext}(a). For some authors the texts are clearly grouped and separated from the rest (e.g. texts from A. C. Doyle and B. Shaw) while for other authors the separation is not as clear. A common trend exists nevertheless, with texts by the same author located in preferential regions in the attribute space.

Even though a direct comparison to related works requires using the same text collection, two examples using collections with similar characteristics which use static network metrics are worth mentioning. A similar study for the same task~\cite{amancio2011comparing} analyzed $40$ texts from $8$ authors in English reaching a success score of $65\%$. In another work, $36$ Persian books from $5$ authors were classified with an accuracy rate of $77.8\%$~\cite{mehri2012complex}. A myriad of other metrics for authorship identification have been proposed. Argamon and Juola~\cite{argamon2011overview} collected the results of the PAN 2011 competition where $3001$ electronic messages from $26$ authors were classified using diverse metrics for which the best micro-averaged recall (i.e. success score) was $0.717$. These collections have characteristics different to ours such as the number of texts, authors, and the sizes of messages compared to books.

\section*{Conclusion}

Network dynamics could be probed in a straightforward manner owing to the stationarity of the series obtained with the network metrics; as a bonus, some of the problems faced in applying networks to real-life problems are solved. Then, texts of different sizes can be compared, and indeed, the smallest book of the collection (from A. C. Doyle) was correctly classified repeatedly. Success scores reached $85\%$, which is outstanding using a collection with such characteristics. Dimensionality reduction through non-linear feature extraction helped to raise the success rates in classification.

Although two of the twelve network metrics did not comply with some stationarity tests, they were successfully used for the task. For instance, transitivity $T$ was largely used in the combinations leading to the highest scores shown in figure~\ref{fig:feat_sel}(d).
The typical sizes of the networks were slightly more than $100$ nodes which are usually considered small. Our approach succeeds because it collects only global metrics, i.e. averages which are still reliable, in contrast to distributions over all nodes. When the typical network sizes are below a few hundreds of nodes, the scores drop. On the other hand, when network sizes are too large the scores also diminish (at a slower pace), because the number of elements in the series decreases as the network size increases.
Considering that small books in the collection are not the source of wrong classification, we conclude that the errors are caused by the variability of style of some authors in their books: while for some authors texts are clearly concentrated in a small region of attribute space, the texts from others are scattered. This reflects that some authors use well-defined structures while others change their narrative resources from one text to another.

Converting networks structure information to time series allows one to use the vast knowledge already applied to other series to analyze evolution of network topologies, and in particular, to the way an author uses the structures offered by the language in his/her narrative.
Purely dynamical measures, i.e. higher moments of the time series, revealed an aspect hitherto unknown of the close relation between style and network dynamics. Because co-occurrence networks are also author-dependent, with network dynamics texts of different sizes can be compared, which further expands the application of networks to real situations.

\section*{acknowledgments}
This work was supported by the Brazilian National Council for Scientific and Technological Development (CNPq). DRA acknowledges financial support from S\~ao Paulo Research Foundation (FAPESP grant no. 14/20830-0).

\end{document}